\documentclass[conference]{IEEEtran}

\usepackage{graphicx}
\usepackage[cmex10]{amsmath}
\usepackage{amsfonts}
\usepackage{amssymb}
\usepackage{xspace}
\usepackage{algorithm,algorithmic}
\usepackage{epstopdf}
\usepackage{diagbox}
\usepackage{pbox}
\newcommand\KM{$K$-means\xspace}

\newcommand\KNN{$K$-nearest neighbors\xspace}

\graphicspath{{./fig/}}

\newcommand{\pic}[4]{
  \begin{figure}[H]
    \centering
    \includegraphics[width=#1\columnwidth]{#2}
    \caption{#3}
    \label{#4}
  \end{figure}
}

\title{Dimensionality reduction for acoustic vehicle classification with spectral embedding}

\author{\IEEEauthorblockN{Justin Sunu}
  \IEEEauthorblockA{Institute of Mathematical Sciences\\
    Claremont Graduate University\\
    Claremont, CA 91711\\
    Email: justinsunu@gmail.com}
  \and
  \IEEEauthorblockN{Allon G.\ Percus}
  \IEEEauthorblockA{Institute of Mathematical Sciences\\
    Claremont Graduate University\\
    Claremont, CA 91711\\
    Email: allon.percus@cgu.edu}}

\begin{document}
\maketitle

\begin{abstract}
We propose a method for recognizing moving vehicles, using data from roadside audio sensors.  This problem has applications ranging widely, from traffic analysis to surveillance.  We extract a frequency signature from the audio signal using a short-time Fourier transform, and treat each time window as an individual data point to be classified.  By applying a spectral embedding, we decrease the dimensionality of the data sufficiently for K-nearest neighbors to provide accurate vehicle identification.
\end{abstract}

\section{Introduction}

Classification and identification of moving vehicles from audio signals is of interest in many applications, ranging from traffic flow management to military target recognition.  Classification may involve differentiating vehicles by type, such as jeep, sedan, etc.  Identification can involve distinguishing specific vehicles, even within a given vehicle type.

Since audio data is small compared to, say, video data, multiple audio sensors can be placed easily and inexpensively.  However, there are certain obstacles having to do with both hardware and physics.  Certain microphones and recording devices have built-in features, for example, damping/normalizing that may be applied when the recording exceeds a threshold.  Microphone sensitivity is another equipment problem: the slightest wind could give disruptive readings that affect the analysis.  Ambient noise is a further issue, adding perturbations to the signal.  Physical challenges include the Doppler shift, where the sound of a vehicle approaching differs from the sound of it leaving, so trying to relate these two can prove difficult.

The short-time Fourier transform (STFT) is often used for feature extraction in audio signals.  We adopt this approach, choosing time windows large enough that they carry sufficient frequency information but small enough that they allow us to localize vehicle events.  Afterwards, we
use spectral embedding as a dimension reduction technique, reducing from thousands of Fourier coefficients to a small number of graph Laplacian eigenvectors. We then cluster the low-dimensional data using \KM, establishing an unsupervised spectral clustering baseline prediction.  Finally, we improve upon this by using \KNN as a simple but highly effective form of semi-supervised learning, giving an accurate classification without the need for large quantities of training data required by frequently used supervised approaches such as deep learning.

In this paper, we apply these methods to audio recordings of passing
vehicles.  In Section 2, we provide background on vehicle classification
using audio signals. In Section 3, we discuss the characteristics of the
vehicle data that we use.  Section 4 describes our feature extraction
methods.  Section 5 discusses our classification methods.  Section 6
presents our results. We conclude in section 7 with a discussion of our method's strengths and limitations, as well as future directions.

\section{Background}

The vast majority of the literature in audio classification is devoted to
speech and music processing, with relatively few papers on problems of
vehicle identification and classification. The most closely related work
has included using principle component analysis for classifying car vs.\
motorcycle~\cite{VSSRbFVPCA}, using an $\epsilon$-neighborhood to cluster
Fourier coefficients to classify different vehicles~\cite{VIuWSN}, and
using both the power spectral density and wavelet transform with $K$-nearest neighbors and support vector machines to classify vehicles \cite{AEoFEMfVCBOAS}. Our study takes a graph-based clustering approach to identifying different individual vehicles from their Fourier coefficients.

Analyzing audio data generally involves the following steps:

\begin{enumerate}
  \item Preprocess raw data.
  \item Extract features in data.
  \item Process extracted data.
  \item Analyze processed data.
\end{enumerate}

The most common form of preprocessing on raw data is ensuring that it has zero mean, by subtracting any bias introduced in the sound recording \cite{VIuWSN,AEoFEMfVCBOAS}.  Another form of preprocessing is applying a weighted window filter to the raw data.  For example, the Hamming window filter is often used to reduce the effects of jump discontinuity when applying the short-time Fourier transform, known as the Gibbs' effect \cite{VSSRbFVPCA}.  The final preprocessing step deals with the  manipulation of data size, namely how to group audio frames into larger windows.  Different window sizes have been used in the literature, with no clear set standard.  Additionally, having some degree of overlap between successive windows can help smooth results \cite{VSSRbFVPCA}.  The basis for these preprocessing steps is to set up the data to allow for better feature extraction.

STFT is frequently used for feature extraction~\cite{VSSRbFVPCA,VIuWSN,AEoFEMfVCBOAS,SFftCoCVuAS}.  Other approaches include the wavelet transform \cite{AEoFEMfVCBOAS,Wbadomv} and the one-third-octave filter bands \cite{MVNCuMC}.  All of these methods aim at extracting underlying information contained within the audio data.

After extracting pertinent features, additional processing is needed.  When working with STFT, the amplitudes for the Fourier coefficients are generally  normalized before analysis is performed~\cite{VSSRbFVPCA,VIuWSN,AEoFEMfVCBOAS,SFftCoCVuAS}.  Another processing step applied to the extracted features is dimension reduction \cite{DimReduction}.  The Fourier transform results in a large number of coefficients, giving a high-dimensional description of the data.  We use a spectral embedding to reduce the dimensionality of the data~\cite{Luxburg2007}.  The spectral embedding requires the use of a distance function on the data points: by adopting the cosine distance, we avoid the need for explicit normalization of the Fourier coefficients.

Finally, the analysis of the processed data involves the classification algorithm.  Methods used for this have included the following:
\begin{itemize}
  \item $K$-means and $K$-nearest neighbors \cite{AEoFEMfVCBOAS}
  \item Support vector machines \cite{AEoFEMfVCBOAS}
  \item Within $\epsilon$ distance \cite{VIuWSN}
  \item Neural networks \cite{MVNCuMC}
\end{itemize}
$K$-means and $K$-nearest neighbors are standard techniques for analyzing
the graph Laplacian eigenvectors resulting from spectral
clustering~\cite{Luxburg2007}.  They are among the simplest methods,
but are also well suited to clustering points in the low-dimensional
space obtained through the dimensionality reduction step.

\section{Data}
Our dataset consists of recordings, provided by the US
Navy's Naval Air Systems Command~\cite{AFPC}, of different vehicles
moving multiple times through a parking lot at approximately 15mph.
While the original dataset consists of MP4 videos taken from a roadside
camera, we extract the dual channel audio signal, and average the
channels together into a single channel.  The audio signal
has a sampling rate of 48,000 frames per second.  Video information is
used to ascertain the ground truth (vehicle identification) for
training data.

The raw audio signal already has zero mean.  Therefore, the only
necessary preprocessing is grouping audio frames into time windows for
STFT.  We found that with windows of $1/8$ of a second, or 6000 frames,
there is both a sufficient number of windows and sufficient information
per window.  This is comparable to window sizes used in other
studies~\cite{VSSRbFVPCA}.

\begin{figure}[]\centering 

\includegraphics[width=\columnwidth]{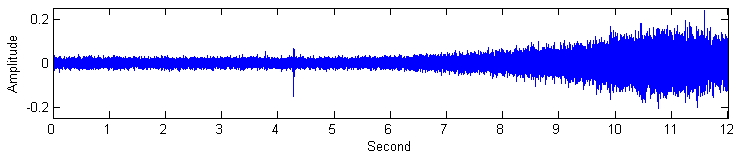}

  \caption{Raw audio signal for single-vehicle data sequence.}
  \label{carloc}
\end{figure}

\begin{figure}[]\centering 
\begin{tabular}{ccc}
  \multicolumn{3}{c}{\includegraphics[width=\columnwidth]{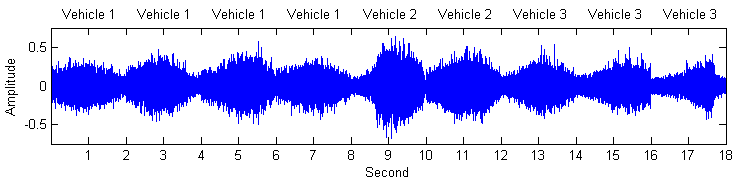}}\\
  \includegraphics[width=.3\columnwidth]{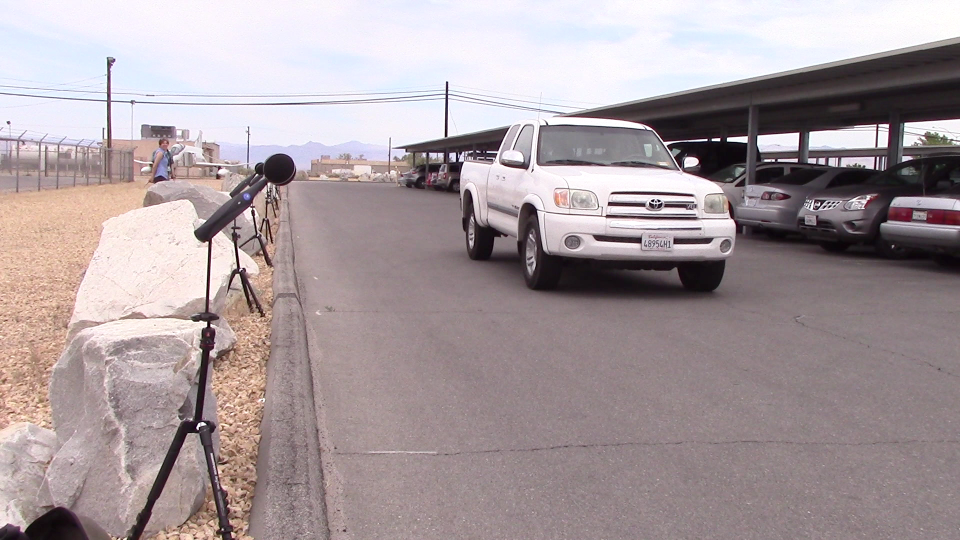} &
  \includegraphics[width=.3\columnwidth]{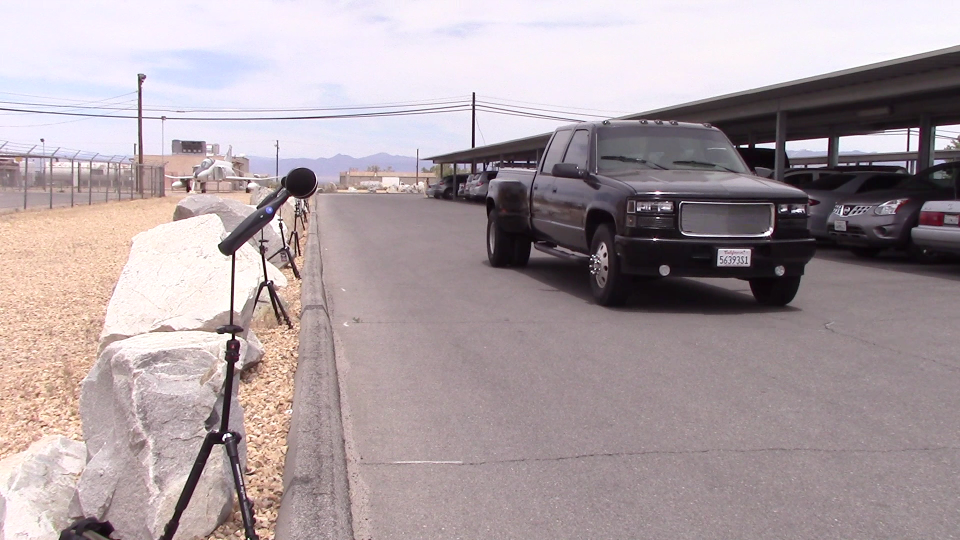} &
  \includegraphics[width=.3\columnwidth]{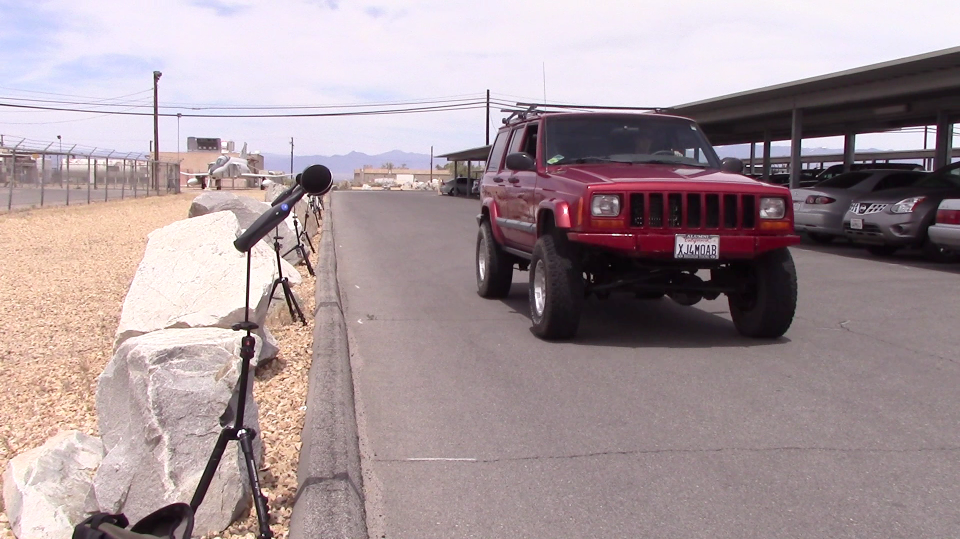} \\
\end{tabular}
  \caption{Raw audio signal for composite data. Images show the three
different vehicles, as seen in accompanying video (not used for
classification).}
  \label{carloc2}
\end{figure}

We use two different types of datasets for our analysis.  The first is a
single audio sequence of a vehicle passing near the microphone, used as
a test case for classifying the different types of sounds involved, differentiating background audio from vehicle audio.
This sequence, whose raw audio signal is shown in Figure~\ref{carloc},
involves the vehicle approaching from a distance, becoming audible after
5 or 6 seconds, passing the microphone after 10 seconds, and then leaving.
The second sequence, shown in Figure~\ref{carloc2}, is a compilation
formed from multiple passages of three different vehicles (a white
truck, black truck, and jeep).  We crop the two seconds where the vehicle is closest to the camera, having the highest amplitude, and then combine these to form a composite signal.  The goal here is test the clustering algorithm's ability to differentiate the vehicles.

\section{Feature extraction}

\subsection{Fourier coefficients}
In order to extract relevant features from our raw audio signals, we use the short-time Fourier transform.

\begin{figure}[]
  \centering
  \includegraphics[width=\columnwidth]{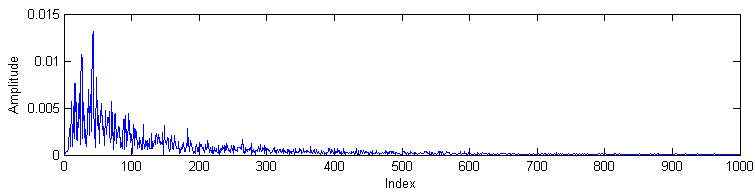}
  \caption{First 1000 Fourier coefficients of a background audio frame}
  \label{f6a}
\end{figure}

With time windows of $1/8$ of a second, or 6000 frames, the Fourier decomposition contains 6000 coefficients.  These are symmetric, leaving 3000 usable coefficients.  Figure \ref{f6a} shows the first 1000 Fourier coefficients for a time window representing background noises.  Note that much of the signal is concentrated within the first 200 coefficients.

\subsection{Fourier reconstructions}

Given the concentration of frequencies, we hypothesize that we can isolate specific sounds by selecting certain ranges of frequency.  To test this, we perform a reconstruction analysis of the Fourier coefficients.  After performing the Fourier transform, we zero out a certain range of Fourier coefficients and then perform the inverse Fourier transform.  This has the effect of filtering out the corresponding range of frequencies.

\begin{figure}[htb]
  \begin{tabular}{c}
    \includegraphics[width=.44\textwidth]{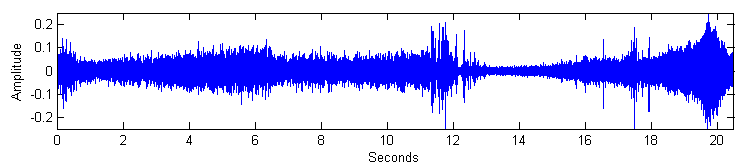} \\ a) Raw signal data.\\
    \includegraphics[width=.44\textwidth]{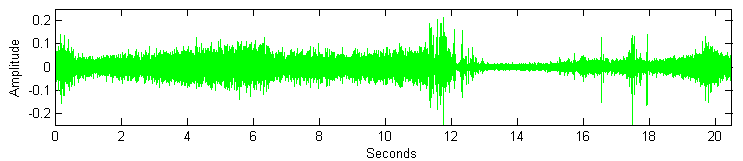} \\ b) Reconstruction using first 130 Fourier coefficients.\\
    \includegraphics[width=.44\textwidth]{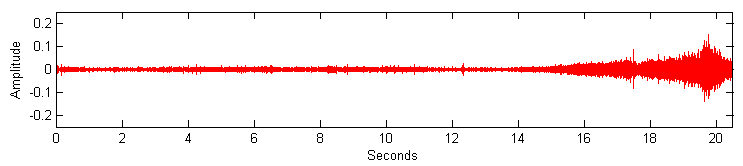} \\ c) Reconstruction using second 130 Fourier coefficients. \\
    \includegraphics[width=.44\textwidth]{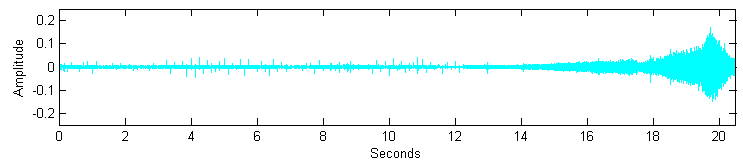} \\ d) Reconstruction using remaining Fourier coefficients.\\
  \end{tabular}
  \caption{Decomposition of an additional (single-vehicle) data sequence into three frequency bands.}
  \label{f10_new}
\end{figure}

Figure \ref{f10_new} shows the results of the reconstruction on an audio recording exhibiting strong wind sounds for the first 12 seconds, before the arrival of the vehicle at second 14.  In a) the raw signal is shown.  In b) we keep only the first 130 coefficients, in c) we keep only the next 130 coefficients, and in d) we keep all the rest of the coefficients.  We see in the reconstruction that the first 130 Fourier coefficients contain most of the background sounds, including the strong wind that corresponds to the large raw signal amplitudes in the first 12 seconds.  The remaining Fourier coefficients are largely insignificant during this time.  When the vehicle becomes audible, however, the second 130 and the rest of the coefficients exhibit a significant increase in amplitude.

By listening to the audio of the reconstructions b) through d), one can
confirm that the first 130 coefficients primarily represent the
background noise, while the second 130 and the rest of the audio capture
most of the sounds of the moving vehicle.
This suggests that further analysis into the detection of background frame signatures could yield a better method for finding which frequencies to 
filter out, in order to yield better reconstructed audio sequences.


\subsection{Vehicle comparisons}
\begin{figure}[]
  \centering
  \includegraphics[width=\columnwidth]{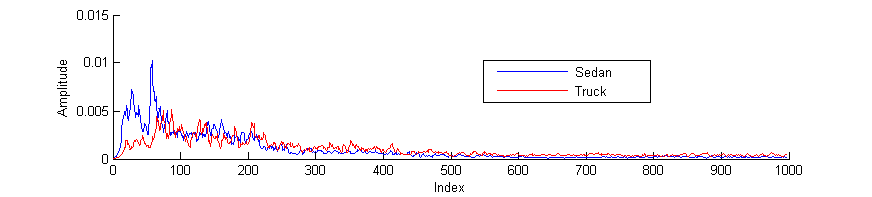}
  \caption{Comparing car and truck Fourier coefficients, after applying a moving mean of size 5.}
  \label{f16_3} 
\end{figure}

The goal of feature extraction is to detect distinguishing characteristics in the data. 
As a further example of why Fourier coefficients form a suitable set of
features for vehicle identification, 
Figure \ref{f16_3} shows Fourier coefficients for a sedan and for a
truck, in both cases for time windows where the vehicle is close to the
microphone.  A moving mean of size 5 is used to smooth the plots, and
coefficients are normalized to sum to 1 in this figure, to enable a
comparison that is affected by different microphone volumes.
There is a clear distinction between the two frequency signatures,
particularly at lower frequencies.  Therefore, in order to distinguish
between different vehicles, we focus on effective ways of clustering
these signatures.


\section{Spectral embedding}
To differentiate background sounds from vehicle sounds, and to identify
individual vehicles, we apply a spectral embedding and then use both $K$-means
and $K$-nearest neighbors as the final classification step.  We treat
each time window as an independent data point to be classified: for
window $i$,
let $\mathbf{x_i}\in\mathbb{R}^m$ represent the set of $m$ Fourier coefficients
associated with that window.

A spectral embedding requires a distance function for comparing the
different data points.  Given that we use a large number of Fourier
coefficients (dimensionality $m$), many of which may be relatively
insignificant, we use the cosine distance so as to decrease our
sensitivity to these small coefficient values.
The distance is given by
$$d_{ij} = 1-\frac{\mathbf{x_i}\cdot
\mathbf{x_j}}{\|\mathbf{x_i}\| ~ \|\mathbf{x_j}\|}$$

%

The goal of a spectral embedding is to reduce the dimensionality of the
data coming from our feature extraction step.
The method, described in Algorithm~\ref{algsc}, involves associating data points
with vertices on a graph, and similarities (Gaussian function of
distance) with edge weights.  By
considering only the leading eigenvectors of the graph Laplacian, we
obtain a description of the data points in a low-dimensional Euclidean
space, where they may be more easily clustered.  This approach allows for greater control than other dimensionality reduction methods for vehicle audio recognition, such as PCA~\cite{VSSRbFVPCA}.
Note that our algorithm uses
an adaptive expression for the Gaussian similarities, with the variance
set according the distance to a point's $N$th neighbor, where $N$ is a
specified parameter.  We also set all similarities beyond the $N$th
neighbor to be zero, to generate a sparser and more easily clustered
graph.

\begin{algorithm}[H] \caption{Spectral embedding pseudo code}\label{algsc}
  \begin{algorithmic}[1]
    \STATE \textbf{INPUT} $n$ data points (STFT coefficients $\mathbf{x_1},\dots,\mathbf{x_n}$)
    \STATE Form distance matrix containing cosine distances between data points, $d_{ij}:i,j=1,\dots,n$ 
    \STATE Compute the Gaussian similarity from each distance, $S_{ij}= e^{-d_{ij}^2/\sigma_i^2}$ , where $\sigma_i$ is the $N$th largest similarity for point $i$
    \STATE Keep only the $N$ largest similarities $S_{ij}$ for every value of $i$, setting all others to zero
    \STATE Form the diagonal matrix, $D_{ii} = \sum_{j}S_{ij}$
    \STATE Symmetric normalized graph Laplacian (SNGL) is defined as
$\mathbf{L_s} = \mathbf{I} - \mathbf{D}^{-1/2} \mathbf{S}
\mathbf{D}^{-1/2}$ 
    \STATE \textbf{OUTPUT} Eigenpairs of the SNGL: eigenvectors $V$ and eigenvalues $\lambda$ 
  \end{algorithmic}
\end{algorithm}

\section{Results}

We used the following parameters and settings:


\begin{itemize}
  \item 6000 frames per time window, resulting in 6000 possible Fourier
coefficients.  We use only the first $m=1500$ coefficients for spectral
clustering, since these coefficients represent 98\% of our data.
  \item Standard box windows, with no overlap.  We found no conclusive
benefit when introducing weighted window filters or overlapping time windows.
  \item $N=15$ for spectral embedding: each node in the graph has
15 neighbors, and the distance to the 15th neighbor is used to establish
the variance in the Gaussian similarity.
\end{itemize}

\subsection{Eigenvectors}

Figure \ref{fsc31} shows the eigenvalues of the Laplacian resulting from the
spectral embedding of the composite 
(multiple-vehicle) dataset.  This spectrum shows gaps after the third
eigenvalue and the fifth eigenvalue, suggesting that a sufficient
number of eigenvectors to use may be either three or
five~\cite{Luxburg2007}.  In practice, we find that five eigenvectors
give good $K$-means and $K$-nearest neighbor clustering results.

\pic{.8}{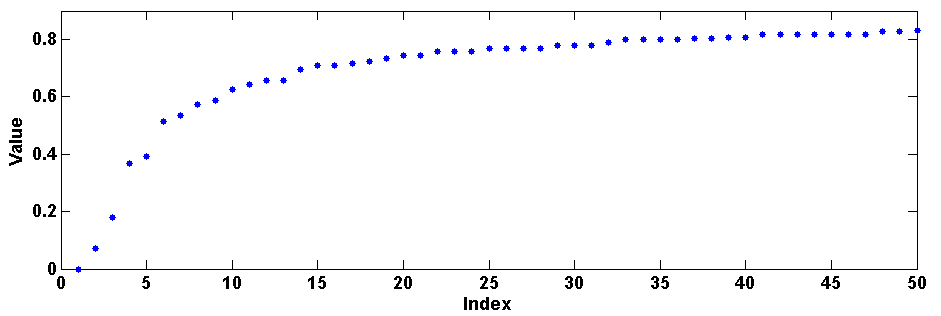}{Spectrum of SNGL for composite dataset, with
$N=15$.  Note gaps after third and fifth eigenvalue.}{fsc31}

\subsection{Spectral clustering}

The spectral clustering method applies $K$-means to the  leading eigenvectors of the
Laplacian.  Figure \ref{fsc12} shows results for the single-vehicle
data, for $K=2,3,4$.  We show ``best''
results over 1000 randomized initializations: we select the
clustering result with the smallest sum of squared distances between
data points and associated cluster centers.  $K=3$ gives a
relatively clear separation of the signal into background noise,
approaching vehicle, and departing vehicle (the latter two sounds
differentiated by Doppler shift). $K=2$ and $K=4$ are less
satisfactory, either clustering the vehicle approach together with
background or subdividing the background cluster.

Figure \ref{fsc32} shows the results of $K$-means on the composite data.  
Given that there are 3 distinct vehicles, $K=3$ is chosen in an attempt
to classify these, and is also consistent with the largest eigengap in
Figure \ref{fsc31} falling after the third eigenvalue.
While $K$-means accurately clusters the majority of the data, many
individual data points are misclassified.  To improve these results, we
instead turn to a semi-supervised classification method.

\pic{.9}{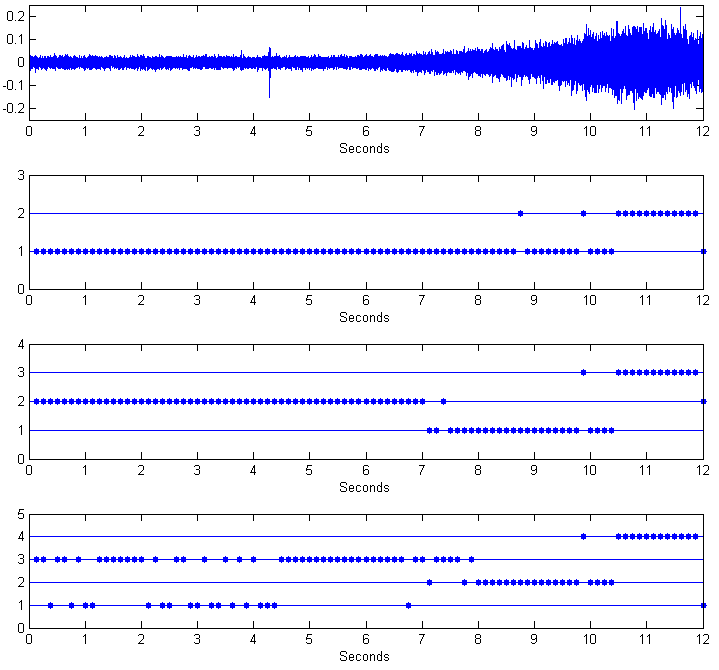}{\KM on eigenvectors from spectral
clustering of single-vehicle data. Raw signal, followed by results for $K=2,3,4$. }{fsc12}

\pic{.9}{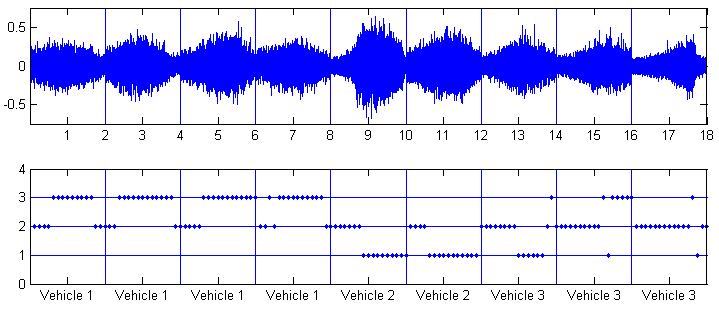}{\KM on eigenvectors from spectral
clustering of composite (multiple-vehicle) data.  Raw signal, followed by results for $K=3$.}{fsc32}

\subsection{Spectral embedding with $K$-nearest neighbors}

We now test $K$-nearest neighbors on the eigenvectors for the composite
data.  We use this as a semi-supervised classification method, training
using one entire audio sample from each of the three different vehicles.
In this way, 
our method reflects an actual application where we might have known
samples of vehicles.  The results for $K=16$ are shown in Figure \ref{fsc34}.
The corresponding confusion matrix is given in Table~\ref{tab2}.  
We allow for training points to be classified outside
of their own class (seen in the case of vehicle 3),
allowing for a better evaluation of the method's accuracy.
While a few data points are misclassified, the vast
majority (88.2\%) are correct.  Training on an entire
vehicle passage appears sufficient to overcome Doppler shift effects in
our data: the approaching sounds and departing sounds of a given vehicle
are correctly placed in the same class.

\pic{.9}{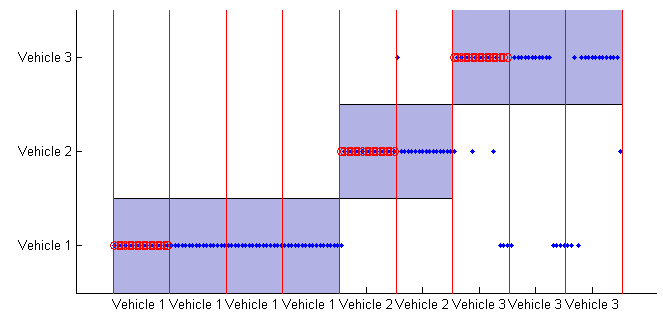}{$K$-nearest neighbors on eigenvectors from
spectral embedding of composite (multiple-vehicle) data, for $K=15$.
Training points are shown with red circles.
Shaded regions show correct classification.}{fsc34}


\begin{table}[h]
  \caption{Classification results for $K$-nearest neighbor.}
  \label{tab2} \centering
  \begin{tabular}{|c|c|c|c|}
    \hline
    \backslashbox{True}{Obtained} & \pbox{2in}{Vehicle 1 \\ \centering (white
truck)} & \pbox{2in}{Vehicle 2 \\ \centering (black truck)} & \pbox{2in}{Vehicle 3
\\ \centering (jeep)} \\
\hline
    Vehicle 1 (white truck)   &     64      &      0      &  0   \\ \hline
    Vehicle 2 (black truck)   &      1      &     30      &  1   \\ \hline
    Vehicle 3 (jeep)       &      11      &      4      &  33  \\ \hline
  \end{tabular}
\end{table}

\section{Conclusions}

Identifying moving vehicles from audio recordings is a challenging and broadly applicable problem.  We have demonstrated an approach that classifies frequency signatures, applying the short-time Fourier transform (STFT) to the audio signal and describing the sound at each $1/8$-second time window using 1500 Fourier coefficients.  Using a spectral embedding, we reduce the dimensionality of the data from 1500 to 5, corresponding to the five eigenvectors of the graph Laplacian.  $K$-nearest neighbors then associates vehicle sounds with the correct vehicle in 88.2\% of the time windows in our test data.

Our analysis treats time windows as independent data points, and therefore ignores temporal correlations.  It is possible that we could improve results by explicitly incorporating time information into our classification algorithm.  For instance, one straightforward approach could be to use as data points a sliding window of larger width.  In some cases, however, ignoring time information could actually help our method, for instance by helping the classifier correctly associate the Doppler-shifted sounds of a given vehicle approaching and departing.

A limitation of our study is that our audio samples only involve single vehicles, under relatively tightly controlled conditions.  The presence of multiple vehicles, or significant external noise such as in an urban environment, would pose a challenge to our feature extraction method.
While the STFT is standard in audio processing, the use of time windows imposes a specific time scale that may not always be appropriate.  Furthermore, the Fourier decomposition may be insufficiently sparse, with too many distinct Fourier components present in vehicle audio signals.  To overcome these difficulties, one could use multiscale techniques such as wavelet decompositions that have been proposed for vehicle detection and classification~\cite{AEoFEMfVCBOAS,Wbadomv}.  More recently developed sparse decomposition methods may also be of use, as they implicitly learn a good choice of basis functions from the data~\cite{SST,HouShi,EWT,ChuiMhaskar}. 


An additional area for improvement is our clustering algorithm.  More sophisticated methods than $K$-means and $K$-nearest neighbors may allow for vehicle identification under less tightly controlled conditions than those in our experiments, or possibly for identifying broad types of vehicles such as cars or trucks.  Such semi-supervised methods would preserve the chief benefit of our approach, namely its applicability in cases where only very limited training data are available.

%
%
%
%
%


\bibliographystyle{IEEEtran}
\bibliography{biblio}

\end{document}